\documentclass[10pt,twocolumn,letterpaper]{article}

\usepackage{cvpr}
\usepackage{times}
\usepackage{epsfig}
\usepackage{graphicx}
\usepackage{amsmath}
\usepackage{amssymb}

\usepackage[pagebackref=true,breaklinks=true,letterpaper=true,colorlinks,bookmarks=false]{hyperref}

\cvprfinalcopy 

\def\cvprPaperID{3379} 

\ifcvprfinal\fi
\begin{document}
\pagenumbering{gobble}

\title{Object-centric Auto-encoders and Dummy Anomalies\\
for Abnormal Event Detection in Video\vspace*{-0.3cm}}

\author{Radu Tudor Ionescu$^{1,2,3}$, Fahad Shahbaz Khan$^{1}$, Mariana-Iuliana Georgescu$^{2,3}$, Ling Shao$^{1}$\\
$^1$Inception Institute of Artificial Intelligence (IIAI), Abu Dhabi, UAE\\
$^2$University of Bucharest, 14 Academiei, Bucharest, Romania\\
$^3$SecurifAI, 21 Mircea Vod\u{a}, Bucharest, Romania\vspace*{-0.3cm}
}

\maketitle

\begin{abstract}
\vspace*{-0.35cm}
Abnormal event detection in video is a challenging vision problem. Most existing approaches formulate abnormal event detection as an outlier detection task, due to the scarcity of anomalous data during training. Because of the lack of prior information regarding abnormal events, these methods are not fully-equipped to differentiate between normal and abnormal events. In this work, we formalize abnormal event detection as a one-versus-rest binary classification problem. Our contribution is two-fold. First, we introduce an unsupervised feature learning framework based on object-centric convolutional auto-encoders to encode both motion and appearance information. Second, we propose a supervised classification approach based on clustering the training samples into normality clusters. A one-versus-rest abnormal event classifier is then employed to separate each normality cluster from the rest. For the purpose of training the classifier, the other clusters act as dummy anomalies. During inference, an object is labeled as abnormal if the highest classification score assigned by the one-versus-rest classifiers is negative. Comprehensive experiments are performed on four benchmarks: Avenue, ShanghaiTech, UCSD and UMN. Our approach provides superior results on all four data sets. On the large-scale ShanghaiTech data set, our method provides an absolute gain of $8.4\%$ in terms of frame-level AUC compared to the state-of-the-art method~\cite{Sultani-CVPR-2018}.
\end{abstract}


\vspace*{-0.4cm}
\section{Introduction}
\vspace*{-0.1cm}

\begin{figure*}[!t]

\begin{center}
\includegraphics[width=0.897\linewidth]{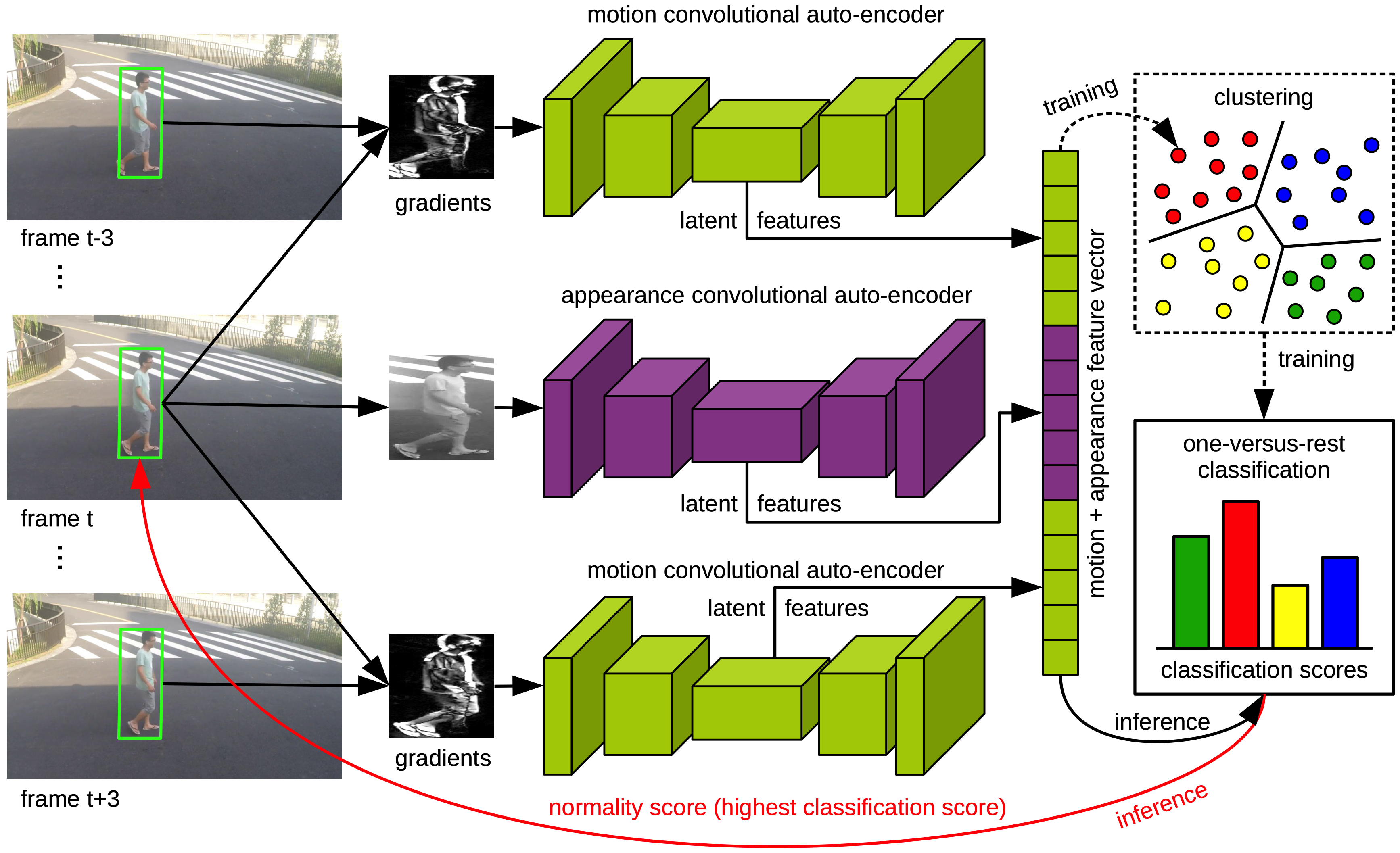}
\end{center}
\vspace*{-0.45cm}
\caption{Our anomaly detection framework based on training convolutional auto-encoders on top of object detections. In the training phase (represented in dashed lines), the concatenated motion and appearance latent representations are clustered and a one-versus-rest classifier is trained to discriminate between the formed clusters. In the inference phase, we label a test sample as abnormal if the highest classification score is negative, i.e. the sample is not attributed to any class. Best viewed in color.}
\label{fig_pipeline}
\vspace*{-0.42cm}
\end{figure*}

Abnormal event detection in video has drawn a lot of attention in the past couple of years~\cite{Giorno-ECCV-2016,Hasan-CVPR-2016,Hinami-ICCV-2017,Ionescu-ICCV-2017,Ionescu-WACV-2019,Liu-CVPR-2018,Liu-BMVC-2018,Luo-ICCV-2017,Ravanbakhsh-WACV-2018,Ravanbakhsh-ICIP-2017,Sabokrou-IP-2017,Smeureanu-ICIAP-2017,Sultani-CVPR-2018,Xu-BMVC-2015,Xu-CVIU-2017,Zhang-PR-2016}, perhaps because it is considered a challenging task due to the commonly accepted definition of abnormal events, which relies on context. An example that illustrates the importance of context is a scenario in which a truck is being driven on the street (normal event) versus a scenario in which a truck is being driven in a pedestrian area (abnormal event). In addition to the reliance on context, abnormal events rarely occur and are generally dominated by more familiar (normal) events.
Therefore, it is difficult to obtain a sufficiently representative set of anomalies, making it hard to employ traditional supervised learning methods.

Most existing anomaly detection approaches~\cite{Antic-ICCV-2011,Cheng-CVPR-2015,Kim-CVPR-2009,Li-PAMI-2014,Lu-ICCV-2013,Mahadevan-CVPR-2010,Mehran-CVPR-2009,Xu-CVIU-2017,Zhao-CVPR-2011} are based on outlier detection and learn a model of normality from training videos containing only familiar events. During inference, events are labeled as abnormal if they deviate from the normality model. Different from these approaches, we address abnormal event detection by formulating the task as a multi-class classification problem instead of an outlier detection problem. Since the training data contains only normal events, we first apply k-means clustering in order to find clusters representing various types of normality (see Figure~\ref{fig_pipeline}). Next, we train a binary classifier following the one-versus-rest scheme in order to separate each normality cluster from the others. During training, normality clusters are treated as different categories, leading to the synthetic generation of abnormal training data.
During inference, the highest classification score corresponding to a given test sample represents the normality score of the respective sample. If the score is negative, the sample is labeled as abnormal (since it does not belong to any normality class). To our knowledge, we are the first to treat the abnormal event detection task as a discriminative multi-class classification problem.

In general, existing abnormal event detection frameworks extract features at a local level~\cite{Giorno-ECCV-2016,Dutta-AAAI-2015,Kim-CVPR-2009,Liu-BMVC-2018,Lu-ICCV-2013,Luo-ICCV-2017,Mahadevan-CVPR-2010,Sabokrou-IP-2017,Saligrama-CVPR-2012,Zhang-PR-2016}, global (frame) level~\cite{Liu-CVPR-2018,Mehran-CVPR-2009,Ravanbakhsh-WACV-2018,Ravanbakhsh-ICIP-2017,Smeureanu-ICIAP-2017}, or both~\cite{Cheng-CVPR-2015,Cong-CVPR-2011,Hasan-CVPR-2016}. All these approaches extract features without explicitly taking into account the objects of interest. In this paper, we propose an object-centric approach by applying a fast yet powerful single-shot detector (SSD)~\cite{Lin-CVPR-2017} on each frame, and learning deep unsupervised features using convolutional auto-encoders (CAE) on top of the detected objects, as shown in Figure~\ref{fig_pipeline}. This enables us to explicity focus only on the objects present in the scene. In addition, it allows us to accurately localize the anomalies in each frame. Although auto-encoders have been used before for abnormal event detection~\cite{Hasan-CVPR-2016,Sabokrou-IP-2017,Xu-CVIU-2017}, to our knowledge, we are the first to train object-centric auto-encoders.

In summary, the novelty of our paper is two-fold. First, we train object-centric convolutional auto-encoders for both motion and appearance. Second, we propose a supervised learning approach by formulating the abnormal event detection task as a multi-class problem. We conduct experiments on the Avenue~\cite{Lu-ICCV-2013}, the ShanghaiTech~\cite{Luo-ICCV-2017}, the UCSD~\cite{Mahadevan-CVPR-2010} and the UMN~\cite{Mehran-CVPR-2009} data sets, and compare our approach with the state-of-the-art abnormal event detection methods~\cite{Cong-CVPR-2011,Giorno-ECCV-2016,Dutta-AAAI-2015,Hasan-CVPR-2016,Hinami-ICCV-2017,Ionescu-ICCV-2017,Ionescu-WACV-2019,Kim-CVPR-2009,Liu-CVPR-2018,Liu-BMVC-2018,Lu-ICCV-2013,Luo-ICCV-2017,Mahadevan-CVPR-2010,Mehran-CVPR-2009,Ravanbakhsh-WACV-2018,Ravanbakhsh-ICIP-2017,Sabokrou-IP-2017,Saligrama-CVPR-2012,Smeureanu-ICIAP-2017,Sultani-CVPR-2018,Xu-BMVC-2015,Xu-CVIU-2017,Zhang-PR-2016}. The empirical results clearly show that our approach achieves superior performance compared to the state-of-the-art methods on all data sets. Furthermore, on the Avenue and the ShanghaiTech data sets, our approach provides considerable absolute gains of $1.5\%$ and $8.4\%$, respectively, over the state-of-the-art methods~\cite{Ionescu-WACV-2019,Sultani-CVPR-2018}. 

We organize the paper as follows. We present related work on abnormal event detection in Section~\ref{sec_RelatedWork}. We describe our approach in Section~\ref{sec_Method}. We present the abnormal event detection experiments in Section~\ref{sec_Experiments}. We draw our final conclusions in Section~\ref{sec_Conclusion}.

\vspace*{-0.2cm}
\section{Related Work}
\label{sec_RelatedWork}
\vspace*{-0.1cm}

Abnormal event detection is commonly formalized as an outlier detection task~\cite{Antic-ICCV-2011,Cheng-CVPR-2015,Cong-CVPR-2011,Dutta-AAAI-2015,Ionescu-WACV-2019,Kim-CVPR-2009,Li-PAMI-2014,Lu-ICCV-2013,Mahadevan-CVPR-2010,Mehran-CVPR-2009,Ren-BMVC-2015,Xu-BMVC-2015,Xu-CVIU-2017,Zhang-PR-2016,Zhao-CVPR-2011}, in which the main approach is to learn a model of familiarity from training videos and label the detected outliers as abnormal. Several abnormal event detection approaches~\cite{Cheng-CVPR-2015,Cong-CVPR-2011,Dutta-AAAI-2015,Lu-ICCV-2013,Ren-BMVC-2015} learn a dictionary of atoms representing normal events during training, then label the events not represented in the dictionary as abnormal. Some recent approaches have employed locality sensitive hashing~\cite{Zhang-PR-2016} and deep learning~\cite{Hasan-CVPR-2016,Hinami-ICCV-2017,Liu-CVPR-2018,Luo-ICCV-2017,Ravanbakhsh-WACV-2018,Ravanbakhsh-ICIP-2017,Sabokrou-IP-2017,Smeureanu-ICIAP-2017,Xu-BMVC-2015,Xu-CVIU-2017} to achieve better results. For instance, Smeureanu et al.~\cite{Smeureanu-ICIAP-2017} employed a one-class Support Vector Machines (SVM) model based on deep features provided by convolutional neural networks (CNN) pre-trained on the ILSVRC benchmark~\cite{Russakovsky2015}, while Ravanbakhsh et al.~\cite{Ravanbakhsh-WACV-2018} combined pre-trained CNN models with low-level optical-flow maps. 

Similar to our own approach, which learns features in an unsupervised fashion, there are a few works that have employed unsupervised steps for abnormal event detection~\cite{Dutta-AAAI-2015,Hasan-CVPR-2016,Ren-BMVC-2015,Sabokrou-IP-2017,Xu-BMVC-2015,Xu-CVIU-2017}. 
Interestingly, some recent works do not require training data at all, in order to detect abnormal events~\cite{Giorno-ECCV-2016,Ionescu-ICCV-2017,Liu-BMVC-2018}. 
More closely-related to our work are methods that employ features learned with auto-encoders~\cite{Hasan-CVPR-2016,Sabokrou-IP-2017,Xu-BMVC-2015,Xu-CVIU-2017} or extracted from the classification branch of Fast R-CNN~\cite{Hinami-ICCV-2017}. In order to learn deep features without supervision, Xu et al.~\cite{Xu-BMVC-2015,Xu-CVIU-2017} used Stacked Denoising Auto-Encoders on multi-scale patches. To detect abnormal events, Xu et al.~\cite{Xu-BMVC-2015,Xu-CVIU-2017} used one-class SVM on top of the deep features. Hasan et al.~\cite{Hasan-CVPR-2016} employed two auto-encoders, one that is learned on conventional handcrafted features, and another one that is learned in an end-to-end fashion using a fully convolutional feed-forward network. On the other hand, Sabokrou et al.~\cite{Sabokrou-IP-2017} combined 3D deep auto-encoders and 3D convolutional neural networks into a cascaded framework.

\noindent
{\bf Differences of our approach.}
Different from these recent related works~\cite{Hasan-CVPR-2016,Sabokrou-IP-2017,Xu-BMVC-2015,Xu-CVIU-2017}, we propose to train auto-encoders on object detections provided by a state-of-the-art detector~\cite{Lin-CVPR-2017}. The most similar work to ours is that of Hinami et al.~\cite{Hinami-ICCV-2017}. They also proposed an object-centric approach, but our detection, feature extraction and training stages are different. While Hinami et al.~\cite{Hinami-ICCV-2017} used geodesic~\cite{Koltun-ECCV-2014} and moving object proposals~\cite{Fragkiadaki-ICCV-2015}, we employ a single-shot detector~\cite{Lin-CVPR-2017} based on Feature Pyramid Networks (FPN). In the feature extraction stage, Hinami et al.~\cite{Hinami-ICCV-2017} fine-tuned the classification branch of the Fast R-CNN model on multiple visual tasks to exploit semantic information that is useful for detecting and recounting abnormal events. In contrast, we learn unsupervised deep features with convolutional auto-encoders. Also differing from Hinami et al.~\cite{Hinami-ICCV-2017} and all other works, we formalize the abnormal event detection task as a multi-class problem and propose to train a one-versus-rest SVM on top of k-means clusters. A similar approach was adopted by Caron et al.~\cite{Caron-ECCV-2018} in order to train deep generic visual features in an unsupervised manner.

\vspace*{-0.2cm}
\section{Method}
\label{sec_Method}
\vspace*{-0.1cm}

\noindent
{\bf Motivation.}
Since the training data contains only normal events, supervised learning methods that require both positive (normal) and negative (abnormal) samples cannot be directly applied for the abnormal event detection task. However, we believe that including any form of  supervision is an important step towards obtaining better performance in practice. Motivated by this intuition, we conceive a framework that incorporates two approaches for including supervision. The first approach consists of employing a single-shot object detector~\cite{Lin-CVPR-2017}, which is trained in a supervised fashion, in order to obtain object detections that are subsequently used throughout the rest of the processing pipeline. The second approach consists of training supervised one-versus-rest classifiers on artificially-generated classes representing different kinds of normality. The classes are generated by previously clustering the training samples. Our entire framework is composed of four sequential stages that are described in detail below. These are the object detection stage, the feature learning stage, the model training stage, and the inference stage.

\noindent
{\bf Object detection.}
We propose to detect objects using a single-shot object detector based on FPN~\cite{Lin-CVPR-2017}, which offers an optimal trade-off between accuracy and speed. This object detector is specifically chosen because $(i)$ it can accurately detect smaller objects, due to the FPN architecture, and $(ii)$ it can process about $13$ frames per second on a GPU. These advantages are of utter importance for developing a practical abnormal event detection framework. The object detector is applied on a frame by frame basis in order to obtain a set of bounding boxes for the objects in each frame $t$. We use the bounding boxes to crop the objects. The resulting images are converted to grayscale. Next, the images are directly passed to the feature learning stage, in order to learn object-centric appearance features. At the same time, we use the images containing objects in order to compute gradients representing motion. For this step, we additionally consider the images cropped from a previous and a subsequent frame. As illustrared in Figure~\ref{fig_pipeline}, we choose the frames at index $t-3$ and $t+3$, with respect to the current frame $t$. Since the temporal distance between the frames is not significant, we do not need to track the objects. Instead, we simply consider the bounding boxes determined at frame $t$ in order to crop the objects at frames $t-3$ and $t+3$. For each object, we obtain two image gradients, one representing the change in motion from frame $t-3$ to frame $t$ and one representing the change in motion from frame $t$ to frame $t+3$. Finally, the image gradients are also passed to the feature learning stage, in order to learn object-centric motion features.
 
\begin{figure}[!t]

\begin{center}
\includegraphics[width=1.0\linewidth]{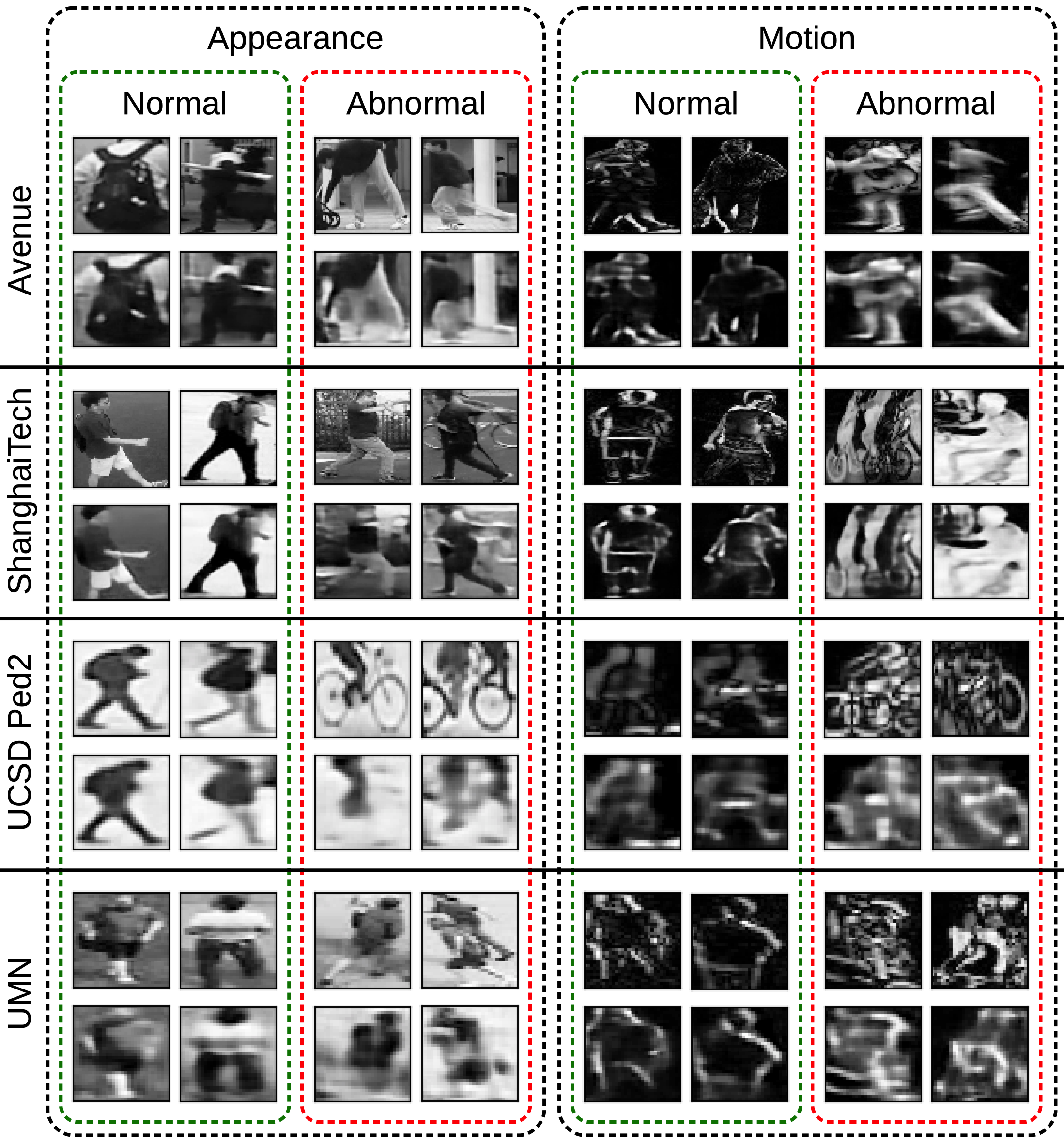}
\end{center}
\vspace*{-0.4cm}
\caption{Normal and abnormal objects (left) and gradients (right) with reconstructions provided by the appearance (left) and the motion (right) convolutional auto-encoders. The samples are selected from the Avenue~\cite{Lu-ICCV-2013}, the ShanghaiTech~\cite{Luo-ICCV-2017}, the UCSD Ped2~\cite{Mahadevan-CVPR-2010} and the UMN~\cite{Mehran-CVPR-2009} test videos, and are not seen during training the auto-encoders.}
\label{fig_cae}
\vspace*{-0.7cm}
\end{figure}

\noindent
{\bf Feature learning.}
In order to obtain a feature vector for each object detection, we train three convolutional auto-encoders. One auto-encoder takes as input cropped images containing objects, and it inherently learns latent appearance features. The other two auto-econders take as input the gradients that capture how the object moved before and after the detection moment, respectively. These auto-encoders learn latent motion features. All three auto-encoders are based on the same lightweight architecture, which is composed of an encoder with $3$ convolutional and max-pooling blocks, and a decoder with $3$ upsampling and convolutional blocks and an additional convolutional layer for the final output. For each CAE, the size of the input is $64 \times 64 \times 1$, and the size of the output is the same. All convolutional layers are based on $3 \times 3$ filters. Each convolutional layer, except the very last one, is followed by ReLU activations. The first two convolutional layers of the encoder contain $32$ filters each, while the third convolutional layer contains $16$ filters. The max-pooling layers of the encoder are based on $2 \times 2$ filters with stride $2$. The resulting latent feature representation of each CAE is composed of $16$ activation maps of size $8 \times 8$. In the decoder, each resize layer upsamples the input activations by a factor of two, using the nearest neighbor approach. The first convolutional layer in the decoder contains $16$ filters. The following two convolutional layers of the decoder contain $32$ filters each. The fourth (and last) convolutional layer of the decoder contains a single filter of size $3 \times 3$. The main purpose of the last convolutional layer is to reduce the output depth from $64 \times 64 \times 32$ to $64 \times 64 \times 1$. The auto-encoders are trained with the Adam optimizer~\cite{Kingma-ICLR-2014} using the pixel-wise mean squared error as loss function:
\begin{equation}\label{eq_2}
\mathcal{L} (I,O) = \frac{1}{h \cdot w}\sum_{i=1}^{h}\sum_{j=1}^{w} \left( I_{ij} - O_{ij}  \right)^2,
\end{equation}
where $I$ and $O$ are the input and the output images, each of size $h \times w$ pixels (in our case, $h=w=64$).

The auto-encoders learn to represent objects detected in the training video containing only normal behavior. When we provide as input objects with abnormal behavior, the reconstruction error of the auto-encoders is expected to be higher.  Furthermore, the latent features should represent known (normal) objects in a different and better way than unknown (abnormal) objects. Some input-output CAE pairs selected from the test videos in each data set considered in the evaluation are shown in Figure~\ref{fig_cae}. We notice that the auto-encoders generally provide better reconstructions for normal objects, confirming our intuition. The final feature vector for each object detection sample is a concatenation of the latent appearance features and the latent motion features. Since the latent activation maps of each CAE are $8 \times 8 \times 16$, the final feature vectors have $3072$ dimensions.

\noindent
{\bf Model training.}
We propose a novel training approach by formalizing the abnormal event detection task as a multi-class classification problem. The proposed approach aims to compensate for the lack of truly abnormal training samples, by constructing a context in which a subset of normal training samples can play the role of \emph{dummy abnormal samples} with respect to another subset of normal training samples. This is achieved by clustering the normal training samples into $k$ clusters using k-means. We consider that each cluster represents a certain kind of normality, different from the other clusters. From the perspective of a given cluster $i$, the samples belonging to the other clusters (from the set $\{1, 2, ...., k\} \setminus i$) can be viewed as (dummy) abnormal samples. Therefore, we can train a binary classifier $g_i$, in our case an SVM, to separate the positively-labeled data points in a cluster $i$ from the negatively-labeled data points in clusters $\{1, 2, ...., k\} \setminus i$, as follows:
\begin{equation}\label{eq_2}
g_i(x) = \sum_{j=1}^{m} w_j \cdot x_j + b,
\end{equation}
where $x \in \mathbb{R}^m$ is a test sample that must be classified either as normal or abnormal, $w$ is the vector of weights and $b$ is the bias term. We note that the negative samples can actually be considered as more closely-related to the samples in cluster $i$ than truly abnormal samples. Hence, the discrimination task is more difficult, and it can help the SVM to select better support vectors. For each cluster $i$, we train an independent binary classifier $g_i$. The final classification score for one data sample is the highest score among the scores returned by the $k$ classifiers. In other words, the classification score for one data sample is selected according to the one-versus-rest scheme, commonly used when binary classifiers are employed for solving multi-class problems.

\noindent
{\bf Inference.}
In the inference phase, each test sample $x$ is classified by the $k$ binary SVM models. The highest classification score is used (with a change of sign) as the abnormality score $s$ for the respective test sample $x$:
\setlength{\abovedisplayskip}{3pt}
\setlength{\belowdisplayskip}{3pt}
\begin{equation}\label{eq_3}
s(x) = - \max\limits_{i} \{ g_i(x) \}, \forall i \in \{1,2,....,k\}.
\end{equation}
By putting together the scores of the objects cropped from a given frame, we obtain a pixel-level anomaly prediction map for the respective frame. If the bounding boxes of two objects overlap, we keep the maximum abnormality score for the overlapping region. To obtain frame-level predictions, we take the highest score in the prediction map as the anomaly score of the respective frame. Finally, we apply a Gaussian filter to temporally smooth the frame-level anomaly scores.

\vspace*{-0.2cm}
\section{Experiments}
\label{sec_Experiments}

\vspace*{-0.1cm}
\subsection{Data Sets}
\vspace*{-0.1cm}


\noindent
{\bf Avenue.}
The Avenue data set~\cite{Lu-ICCV-2013} consists of $16$ training videos with a total $15328$ frames and $21$ test videos with a total of $15324$. The resolution of each video frame is $360 \times 640$ pixels. For each test frame, ground-truth locations of anomalies are provided using pixel-level masks. 

\noindent
{\bf ShanghaiTech.}
The ShanghaiTech Campus data set~\cite{Luo-ICCV-2017} is among the largest data sets for abnormal event detection. Unlike other data sets, it contains $13$ different scenes with various lighting conditions and camera angles. There are $330$ training videos and $107$ test videos. The test set contains a total of $130$ abnormal events annotated at the pixel-level. There are $316154$ frames in the whole data set. The resolution of each video frame is $480 \times 856$ pixels.

\noindent
{\bf UCSD.}
The UCSD Pedestrian data set~\cite{Mahadevan-CVPR-2010} is composed of two subsets, namely Ped1 and Ped2. As Hinami et al.~\cite{Hinami-ICCV-2017}, we exclude Ped1 from the evaluation, because it has a significantly lower frame resolution of $158 \times 238$. Another problem with Ped1 is that some recent works report results only on a subset of $16$ videos~\cite{Ravanbakhsh-WACV-2018,Ravanbakhsh-ICIP-2017,Xu-BMVC-2015}, while others~\cite{Ionescu-ICCV-2017,Mahadevan-CVPR-2010,Liu-CVPR-2018,Liu-BMVC-2018} report results on all $36$ test videos.
We thus consider only UCSD Ped2, which contains $16$ training and $12$ test videos. The resolution of each frame is $240 \times 360$ pixels. There are $2550$ frames for training and $2010$ for testing. 
The videos illustrate various crowded scenes, and anomalies include bicycles, vehicles, skateboarders and wheelchairs crossing pedestrian areas.

\noindent
{\bf UMN.}
The UMN Unusual Crowd Activity data set~\cite{Mehran-CVPR-2009} consists of three independent crowded scenes of different lengths. The three scenes consist of $1453$ frames, $4144$ frames and $2144$ frames, respectively. The resolution of each video frame is $240 \times 320$ pixels. The normal behavior is represented by people walking around, while the abnormal behavior is represented by people running in different directions.

\vspace*{-0.1cm}
\subsection{Evaluation}
\vspace*{-0.1cm}

As evaluation metric, we employ the \emph{area under the curve} (AUC) computed with regard to ground-truth annotations at the frame-level. 
The frame-level AUC metric used in most previous works~\cite{Cong-CVPR-2011,Giorno-ECCV-2016,Ionescu-ICCV-2017,Ionescu-WACV-2019,Liu-CVPR-2018,Liu-BMVC-2018,Lu-ICCV-2013,Luo-ICCV-2017,Mahadevan-CVPR-2010,Sultani-CVPR-2018,Xu-BMVC-2015} considers a frame as being a correct detection, if it contains at least one abnormal pixel.
We adopt the same frame-level AUC definition as these previous works.  
In order to obtain the final abnormality maps, our pixel-level detection maps are smoothed using a similar technique as~\cite{Giorno-ECCV-2016,Ionescu-ICCV-2017,Lu-ICCV-2013}. 

\vspace*{-0.1cm}
\subsection{Parameter and Implementation Details}
\vspace*{-0.1cm}

In the object detection stage, we employ a single-shot detector based on FPN~\cite{Lin-CVPR-2017} that is pre-trained on the COCO data set~\cite{Lin-ECCV-2014}. The detector is downloaded from the TensorFlow detection model zoo. For the training set, we keep the detections with a confidence level higher than $0.5$, and for the test set, we keep those with a confidence level higher than $0.4$. The convolutional auto-encoders used in the feature learning stage are implemented in TensorFlow~\cite{Abadi-OSDI-2016}. We train the auto-encoders for $100$ epochs with the learning rate set to $10^{-3}$, and for another $100$ epochs with the learning rate set to $10^{-4}$. We use mini-batches of $64$ samples. We train independent auto-encoders for each of the four data sets considered in the evaluation. To cluster the training samples with k-means, we employ the VLFeat~\cite{vedaldi-vlfeat-2008} implementation, which is based on the Lloyd algorithm~\cite{Du-SIAM-1999}. We adopt k-means++~\cite{Arthur-SODA-2007} initialization. We repeat the clustering $10$ times, selecting the partitioning with the minimum energy. In all the experiments, we set the number of k-means clusters to $k=10$. We set the regularization parameter of the linear SVM (implemented in VLFeat~\cite{vedaldi-vlfeat-2008}) to $C=1$.

\vspace*{-0.1cm}
\subsection{Results}
\vspace*{-0.1cm}

\begin{table}[t]
\setlength\tabcolsep{4.0pt}
\small{
\begin{center}
\begin{tabular}{|l|c|c|c|c|}
\hline
Method 																& Avenue			& Shanghai		& UCSD  			& UMN \\
			 																&						& Tech				& Ped2 				&  \\
\hline
\hline
Kim et al.~\cite{Kim-CVPR-2009}							& -					& - 					& $69.3$ 			& - \\
Mehran et al.~\cite{Mehran-CVPR-2009}				& -					& - 					& $55.6$ 			& $96.0$ \\
Mahadevan et al.~\cite{Mahadevan-CVPR-2010}	& -					& -	 				& $82.9$	 		& - \\
Cong et al.~\cite{Cong-CVPR-2011}						& -					& - 					& -					& $97.8$ \\
Saligrama et al.~\cite{Saligrama-CVPR-2012}		& -					& -	 				& -					& $98.5$ \\
Lu et al.~\cite{Lu-ICCV-2013}								& $80.9$			& - 					& -					& - \\
Dutta et al.~\cite{Dutta-AAAI-2015}						& -					& - 					& -					& $99.5$ \\
Xu et al.~\cite{Xu-BMVC-2015,Xu-CVIU-2017}		& -					& - 					& $90.8$			& - \\
Hasan et al.~\cite{Hasan-CVPR-2016}					& $70.2$			& $60.9$			& $90.0$			& - \\
Del Giorno et al.~\cite{Giorno-ECCV-2016}			& $78.3$			& -					& - 					& $91.0$ \\
Zhang et al.~\cite{Zhang-PR-2016}						& -					& - 					& $91.0$			& $98.7$ \\
Smeureanu et al.~\cite{Smeureanu-ICIAP-2017}	& $84.6$			& -					& -					& $97.1$ \\
Ionescu et al.~\cite{Ionescu-ICCV-2017}				& $80.6$			& -					& $82.2$			& $95.1$ \\
Luo et al.~\cite{Luo-ICCV-2017}							& $81.7$			& $68.0$ 			& $92.2$			& - \\
Hinami et al.~\cite{Hinami-ICCV-2017}					& - 					& -					& $92.2$			& - \\
Ravanbakhsh et al.~\cite{Ravanbakhsh-ICIP-2017}		& -			& -					& $93.5$ 			& $99.0$ \\
Sabokrou et al.~\cite{Sabokrou-IP-2017}				& -					& - 					& -					& $\mathbf{99.6}$ \\
Ravanbakhsh et al.~\cite{Ravanbakhsh-WACV-2018}	& -			& -					& $88.4$ 			& $98.8$ \\
Liu et al.~\cite{Liu-CVPR-2018}								& $85.1$			& $72.8$ 			& $95.4$			& - \\
Liu et al.~\cite{Liu-BMVC-2018}							& $84.4$			& - 					& $87.5$			& $96.1$ \\
Sultani et al.~\cite{Sultani-CVPR-2018}					& -					& $76.5$			& -					& - \\
Ionescu et al.~\cite{Ionescu-WACV-2019}				& $88.9$			& -					& -					& $99.3$ \\
\hline
Ours															& $\mathbf{90.4}$	& $\mathbf{84.9}$	& $\mathbf{97.8}$	& $\mathbf{99.6}$ \\
\hline
\end{tabular}
\end{center}
\vspace*{-0.15cm}
\caption{Abnormal event detection results (in $\%$) in terms of frame-level AUC on the Avenue~\cite{Lu-ICCV-2013}, the ShanghaiTech~\cite{Luo-ICCV-2017}, the UCSD Ped2~\cite{Mahadevan-CVPR-2010} and the UMN~\cite{Mehran-CVPR-2009} data sets. Our framework is compared with several state-of-the-art approaches~\cite{Cong-CVPR-2011,Giorno-ECCV-2016,Dutta-AAAI-2015,Hasan-CVPR-2016,Hinami-ICCV-2017,Ionescu-ICCV-2017,Ionescu-WACV-2019,Kim-CVPR-2009,Liu-CVPR-2018,Liu-BMVC-2018,Lu-ICCV-2013,Luo-ICCV-2017,Mahadevan-CVPR-2010,Mehran-CVPR-2009,Ravanbakhsh-WACV-2018,Ravanbakhsh-ICIP-2017,Sabokrou-IP-2017,Saligrama-CVPR-2012,Smeureanu-ICIAP-2017,Sultani-CVPR-2018,Xu-BMVC-2015,Xu-CVIU-2017,Zhang-PR-2016}, which are listed in temporal order. The results of Sultani et al.~\cite{Sultani-CVPR-2018} on ShanghaiTech are based on their pre-trained model.\label{tab_results}}
}
\vspace*{-0.4cm}
\end{table}

We evaluate our approach in comparison with a series of state-of-the-art methods~\cite{Cong-CVPR-2011,Giorno-ECCV-2016,Dutta-AAAI-2015,Hasan-CVPR-2016,Hinami-ICCV-2017,Ionescu-ICCV-2017,Kim-CVPR-2009,Liu-CVPR-2018,Liu-BMVC-2018,Lu-ICCV-2013,Luo-ICCV-2017,Mahadevan-CVPR-2010,Mehran-CVPR-2009,Ravanbakhsh-WACV-2018,Ravanbakhsh-ICIP-2017,Sabokrou-IP-2017,Saligrama-CVPR-2012,Smeureanu-ICIAP-2017,Xu-BMVC-2015,Xu-CVIU-2017,Zhang-PR-2016} on the Avenue, the ShanghaiTech, the UCSD Ped2 and the UMN data sets. The corresponding results are presented in Table~\ref{tab_results}. 

\noindent
{\bf Avenue.}
On the Avenue data set, we are able to surpass the results reported in all previous works. Compared to most of the recent works~\cite{Ionescu-ICCV-2017,Liu-CVPR-2018,Liu-BMVC-2018,Luo-ICCV-2017,Smeureanu-ICIAP-2017}, our method provides an absolute gain of more than $5\%$ in terms of frame-level AUC. With a frame-level AUC of $88.9\%$, Ionescu et al.~\cite{Ionescu-WACV-2019} is the best and most recent baseline. We surpass their score by $1.5\%$. Remarkably, with a frame-level AUC of $90.4\%$, our approach is the only method that surpasses the threshold of $90\%$ on the Avenue data set.

Notably, Hinami et al.~\cite{Hinami-ICCV-2017} did not compare with other approaches on the official Avenue test set, arguing that there are five test videos ($01$, $02$, $08$, $09$ and $10$) containing static abnormal objects that are not properly labeled. Therefore, they only evaluated their method on Avenue17, a subset that excludes the respective five videos. We also compare our performance with that reported by Hinami et al.~\cite{Hinami-ICCV-2017}, being sure to exclude the same five test videos for a fair comparison. Our frame-level AUC score on the Avenue17 subset is $91.6\%$, which is almost $2\%$ better than the frame-level AUC of $89.8\%$ reported in~\cite{Hinami-ICCV-2017}. With respect to the complete Avenue test set, we note that our framework attains a better frame-level AUC score on the Avenue17 subset, suggesting that the removed test videos are indeed more problematic than the videos left in Avenue17. As observed by Hinami et al.~\cite{Hinami-ICCV-2017}, the removed videos include some abnormal objects that are not labeled accordingly. Methods detecting these objects as abnormal are destined to reach higher false positive rates, which is unfair.

\begin{figure}[!t]
\begin{center}
\includegraphics[width=0.928\columnwidth]{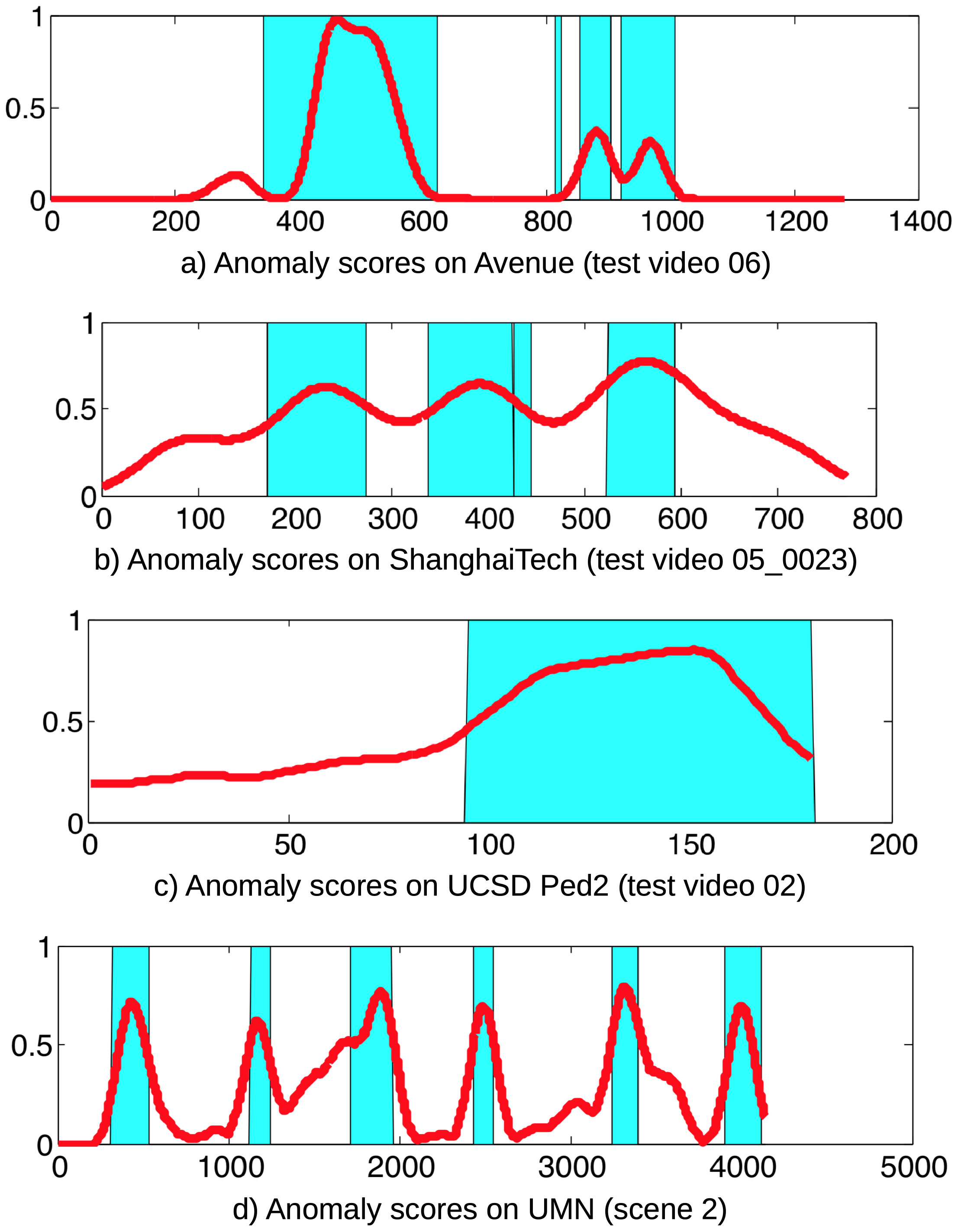}
\end{center}
\vspace*{-0.4cm}
\caption{Frame-level anomaly detection scores between $0$ and $1$ (on the horizontal axis) provided by our approach, for various test videos selected from the Avenue~\cite{Lu-ICCV-2013}, the ShanghaiTech~\cite{Luo-ICCV-2017}, the UCSD Ped2~\cite{Mahadevan-CVPR-2010} and the UMN~\cite{Mehran-CVPR-2009} data sets. Ground-truth abnormal events are represented in cyan and our scores are illustrated in red. Best viewed in color.}
\label{fig_vid_scores}
\vspace*{-0.45cm}
\end{figure}

\begin{figure*}[!t]
\begin{center}
\includegraphics[width=0.852\linewidth]{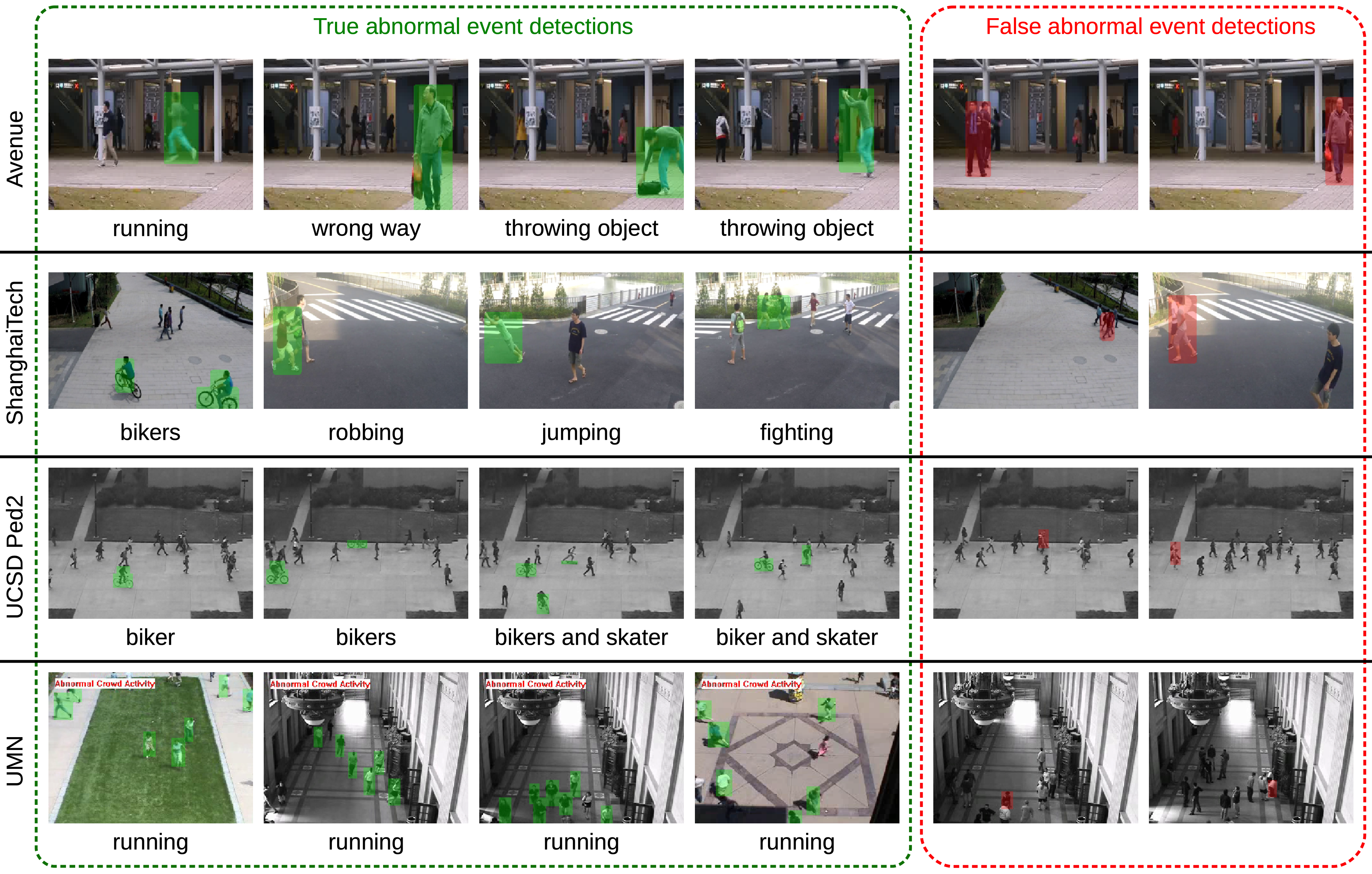}
\end{center}
\vspace*{-0.4cm}
\caption{True positive (left) versus false positive (right) detections of our framework. Examples are selected from the Avenue~\cite{Lu-ICCV-2013} (first row), the ShanghaiTech~\cite{Luo-ICCV-2017} (second row), the UCSD Ped2~\cite{Mahadevan-CVPR-2010} (third row) and the UMN~\cite{Mehran-CVPR-2009} (fourth row) data sets. Best viewed in color.}
\label{fig_pos_neg}
\vspace*{-0.4cm}
\end{figure*}

In Figure~\ref{fig_vid_scores} (a), we present the frame-level anomaly scores provided by our method on test video $06$ from Avenue. According to the ground-truth labels, which are also illustrated in Figure~\ref{fig_vid_scores} (a), we note that there are four abnormal events in the respective test video. Our approach seems to be able to identify three of the four events, without including any false positive detections. Figure~\ref{fig_pos_neg} (top row) illustrates a few examples of true positive and false positive abnormal event detections. From left to right, the true positive detections are \emph{a person running}, \emph{a person walking in the wrong direction}, \emph{a person picking up an object} and \emph{a person throwing an object}. The first false positive example consists of \emph{two people that are detected in the same bounding box} by the object detector. The other false positive detection is \emph{a person walking in the wrong direction} that is labeled as abnormal too soon.

\noindent
{\bf ShanghaiTech.}
Since ShanghaiTech is the newest data set for abnormal event detection, there are only a few recent approaches reporting results on this data set~\cite{Liu-CVPR-2018,Luo-ICCV-2017}. Besides these, Luo et al.~\cite{Luo-ICCV-2017} additionally evaluated a previously published method~\cite{Hasan-CVPR-2016} when they introduced the data set. On the ShanghaiTech data set, the state-of-the-art performance of $72.8\%$ is reported by Liu et al.~\cite{Liu-CVPR-2018}. We outperform their approach by a large margin of $12.1\%$. In order to compare with Sultani et al.~\cite{Sultani-CVPR-2018} in standard formulation of the abnormal event detection task, we used the open source code provided by Sultani et al.~\cite{Sultani-CVPR-2018} to compute their anomaly scores for the large-scale ShanghaiTech data set. As shown in Table~\ref{tab_results}, the approach of Sultani et al.~\cite{Sultani-CVPR-2018} obtains a frame-level AUC of $76.5\%$, outperforming the best existing method~\cite{Liu-CVPR-2018}. Our approach significantly outperforms both Sultani et al.~\cite{Sultani-CVPR-2018} and Liu et al.~\cite{Liu-CVPR-2018}, achieving a frame-level AUC of $84.9\%$. With a frame-level AUC of $84.9\%$, our approach is the only one to surpass the $80\%$ threshold on ShanghaiTech.

In Figure~\ref{fig_vid_scores} (b), we display our frame-level anomaly scores against the ground-truth labels on a ShanghaiTech test video with three abnormal events. On this video, we can clearly observe a strong correlation between our anomaly scores and the ground-truth labels. Some localization results from different scenes in the ShanghaiTech data set are illustrated in the second row of Figure~\ref{fig_pos_neg}. The true positive abnormal events detected by our framework are (from left to right) \emph{two bikers in a pedestrian area}, \emph{a person robbing another person}, \emph{a person jumping} and \emph{two people fighting}. The false positive abnormal events are triggered because, in each case, there are \emph{two people in the same bounding box} and our system labels the unusual appearance and motion generated by the two objects as abnormal.

\noindent
{\bf UCSD Ped2.}
While older approaches~\cite{Kim-CVPR-2009,Mehran-CVPR-2009} report frame-level AUC scores under $70\%$, most  approaches proposed in the last three years~\cite{Hasan-CVPR-2016,Hinami-ICCV-2017,Liu-CVPR-2018,Liu-BMVC-2018,Luo-ICCV-2017,Ravanbakhsh-WACV-2018,Ravanbakhsh-ICIP-2017,Xu-CVIU-2017,Zhang-PR-2016} reach frame-level AUC scores between $87\%$ and $94\%$ on UCSD Ped2. For instance, the frameworks based on auto-encoders~\cite{Hasan-CVPR-2016,Xu-BMVC-2015,Xu-CVIU-2017} attain results of around $90\%$. Liu et al.~\cite{Liu-CVPR-2018} recently outperformed the previous works, reporting a frame-level AUC of $95.4\%$. We further surpass their state-of-the-art result, reaching the top frame-level AUC of $97.8\%$ on UCSD Ped2. Our score is $2.4\%$ above the score reported by Liu et al.~\cite{Liu-CVPR-2018}, $4.3\%$ above the second-best score reported by Ravanbakhsh et al.~\cite{Ravanbakhsh-ICIP-2017}, and more than $7\%$ higher than the scores reported by other frameworks based on auto-encoders~\cite{Hasan-CVPR-2016,Xu-BMVC-2015,Xu-CVIU-2017}.

As for the other data sets, we compare our frame-level anomaly scores against the ground-truth labels on a test video from UCSD Ped2 in Figure~\ref{fig_vid_scores} (c). On this particular video, our frame-level AUC is above $99\%$, indicating that our approach can precisely detect the abnormal event. Furthermore, the qualitative results presented in the third row of Figure~\ref{fig_pos_neg}, show that our approach can also localize the abnormal events from UCSD Ped2. From left to right, the true positive detections are \emph{a biker in a pedestrian area}, \emph{two bikers in a pedestrian area}, \emph{two bikers and a skater in a pedestrian area} and \emph{a biker and a skater in a pedestrian area}. As for ShanghaiTech, the false positive abnormal detections are caused by \emph{two people in the same bounding box}. 

\noindent
{\bf UMN.}
It appears that UMN is the easiest abnormal event detection data set, because almost all works report frame-level AUC scores higher than $95\%$, with some works~\cite{Dutta-AAAI-2015,Ravanbakhsh-ICIP-2017,Sabokrou-IP-2017} even surpassing $99\%$. The top score of $99.6\%$ is reported by Sabokrou et al.~\cite{Sabokrou-IP-2017}, and we reach the same performance on the UMN data set. We note that the second scene seems to be slightly more difficult than the other two scenes, since our frame-level AUC score on this scene is $99.1\%$, while the frame-level AUC scores on the other scenes are $99.9\%$ and $99.8\%$, respectively. For this reason, we choose to illustrate the frame-level anomaly scores against the ground-truth labels for the second scene from UMN in Figure~\ref{fig_vid_scores} (d). Overall, our anomaly scores correlate well with the ground-truth labels, but there are some normal frames with high abnormality scores just before the third abnormal event in the scene. 

In the fourth row of Figure~\ref{fig_pos_neg}, we present some localization results provided by our framework. The true positive examples represent \emph{people running around in different directions}, while the false positive detections are triggered by \emph{two people in the same bounding box} and \emph{a person bending down to pick up an object}. We note that the false positive examples are selected from the second scene, as we did not find false positive detections in the other two scenes. 

\vspace*{-0.1cm}
\subsection{Discussion}
\vspace*{-0.1cm}

While the results presented in Table~\ref{tab_results} show that our approach can outperform the state-of-the-art methods on four evaluation sets, we also aim to address questions about the robustness of our features and parameter choices, and to discuss the running time of our framework.

\begin{figure}[!t]
\begin{center}
\includegraphics[width=0.75\columnwidth]{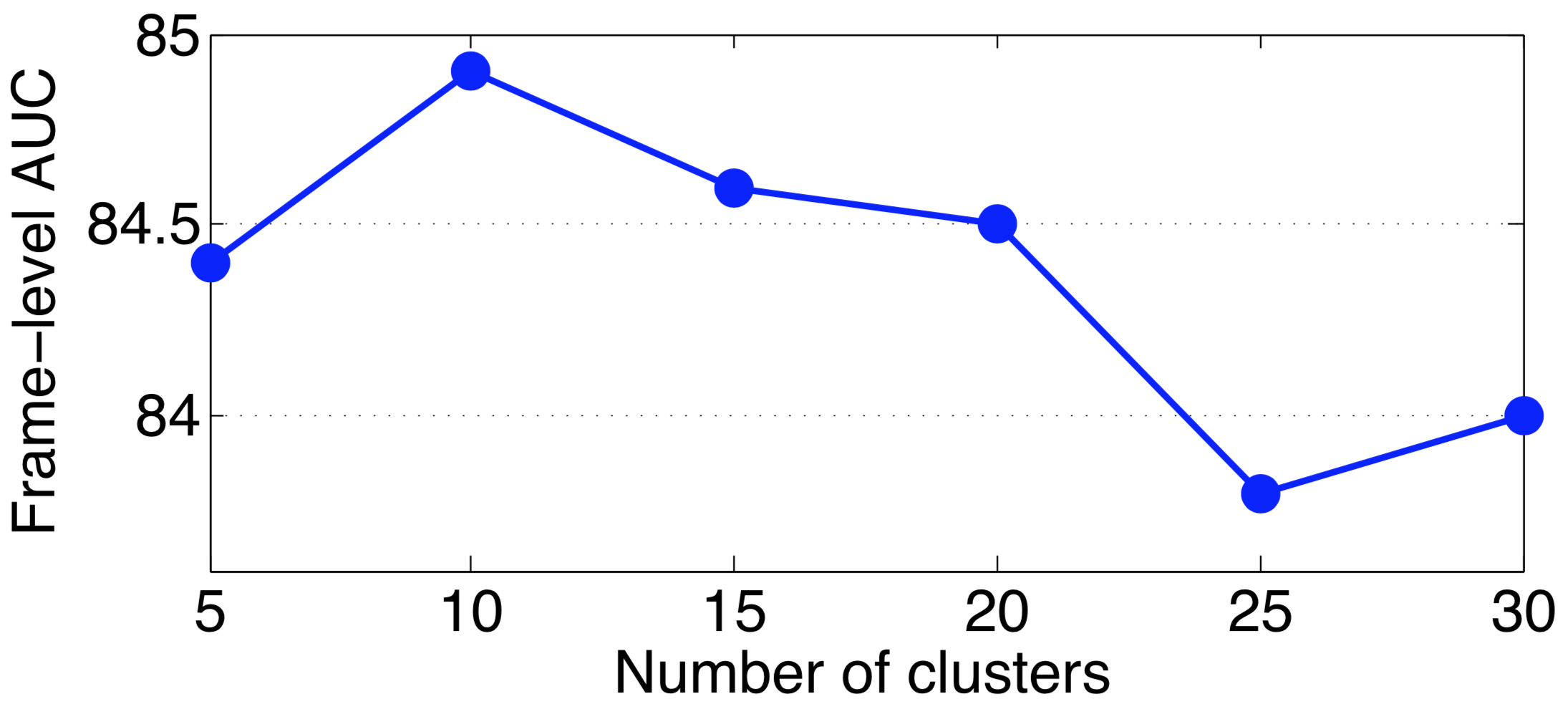}
\end{center}
\vspace*{-0.4cm}
\caption{Frame-level AUC scores on ShanghaiTech obtained by selecting values for the number of clusters $k$ from the set $\{5, 10, 15, 20, 25, 30 \}$.}
\label{fig_var_k}
\vspace*{-0.3cm}
\end{figure}

\begin{figure}[!t]
\begin{center}
\includegraphics[width=0.70\columnwidth]{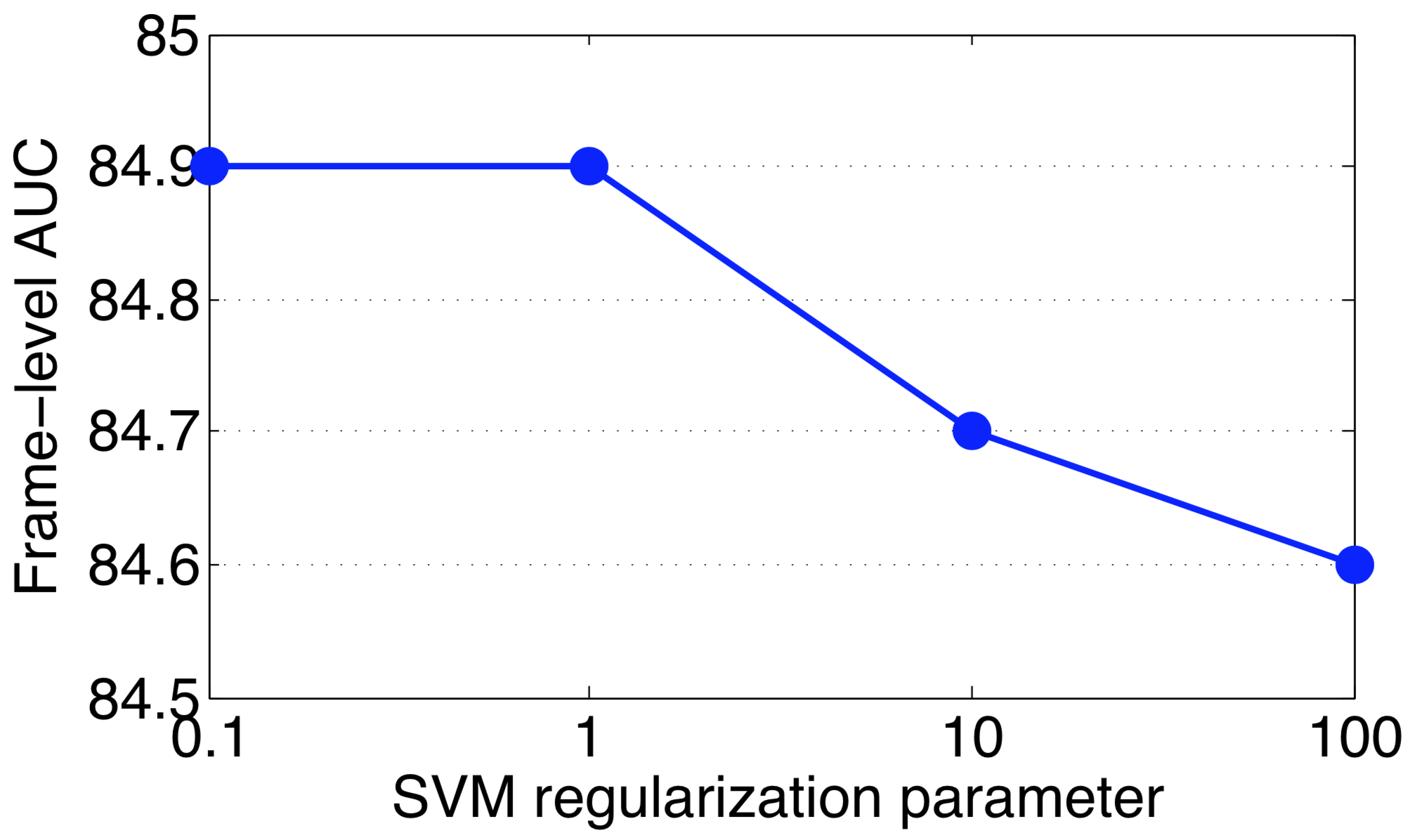}
\end{center}
\vspace*{-0.4cm}
\caption{Frame-level AUC scores on ShanghaiTech obtained by selecting values for the SVM regularization parameter $C$ from the set $\{0.1, 1, 10, 100 \}$.}
\label{fig_var_C}
\vspace*{-0.4cm}
\end{figure}

\noindent
{\bf Parameter selection.}
We present results with various parameter choices on the largest and most difficult evaluation set, namely ShanghaiTech. We first variate the number of clusters $k$ by selecting values in the set $\{5, 10, 15, 20, 25, 30 \}$. The corresponding frame-level AUC scores are presented in Figure~\ref{fig_var_k}. The results presented in Figure~\ref{fig_var_k} indicate that the number of clusters does not play a significant role in our multi-class classification framework, since the accuracy variations are lower than $1.1\%$. With only one exception (for $k=25$), our results are always higher than $84\%$. We also variate the regularization parameter of the SVM, by considering values in the set $\{0.1, 1, 10, 100 \}$. The corresponding frame-level AUC scores are presented in Figure~\ref{fig_var_C}. The results presented in Figure~\ref{fig_var_C} show that the performance variation is lower than $0.3\%$, and the frame-level AUC scores are always higher than $84.6\%$. We believe that this happens because the classes are linearly separable, since they are generated by clustering the samples with k-means into disjoint clusters. Overall, we conclude that our high improvement ($12.1\%$) over the state-of-the-art approach~\cite{Liu-CVPR-2018}, cannot be explained by a convenient choice of parameters.

\noindent
{\bf Ablation results.}
In Table~\ref{tab_ablation}, we present feature ablation results, as well as results for a one-class SVM based on our full object-centric feature set, on the ShanghaiTech data set. When we remove the object detector and train auto-encoders at the frame-level, we obtain a frame-level AUC of $72.4\%$, which demonstrates the importance of extracting object-centric features and using the one-versus-rest SVM. We note that the frame-level auto-encoders have an additional convolutional layer and the input resolution is increased to $192 \times 192$. When we replace the one-class SVM with our multi-class approach based on k-means and one-versus-rest SVM, while keeping the features computed on full frames, the frame-level AUC grows to $78.7\%$. This demonstrates that our approach based on k-means and one-versus-rest SVM is indeed helpful. When we replace the object-centric CAE features with pre-trained SSD features (extracted right before the SSD class predictor), the frame-level AUC is only $81.3\%$, which shows the importance of learning features with auto-encoders. By removing either the appearance or the motion object-centric CAE features from our model, the results drop by less than $3\%$. This shows that both appearance and motion features are relevant for the abnormal event detection task. By replacing our multi-class approach based on k-means and one-versus-rest SVM with a one-class SVM, while keeping the combined object-centric CAE features, the performance drops by $5.7\%$. This result indicates that formalizing the abnormal event detection task as a multi-class problem is indeed useful. We conclude that both of our contributions are crucial to obtain superior results.

\begin{table}[t]
\setlength\tabcolsep{4.0pt}
\small{
\begin{center}
\begin{tabular}{|l|c|}
\hline
Method 																			& Score  \\
\hline
\hline
Frame-level CAE features	+ one-class SVM (baseline)		& $72.4$ \\
\hline
Frame-level CAE features	+ one-versus-rest SVM				& $78.7$ \\
Pre-trained SSD features + one-versus-rest SVM				& $81.3$ \\
CAE appearance features + one-versus-rest SVM				& $82.2$ \\
CAE motion features + one-versus-rest SVM						& $83.0$ \\
Combined CAE features + one-class SVM							& $79.2$ \\
\hline
Combined CAE features + one-versus-rest SVM				& $\mathbf{84.9}$ \\
\hline
\end{tabular}
\end{center}
\vspace*{-0.15cm}
\caption{Frame-level AUC scores (in $\%$) on ShanghaiTech~\cite{Luo-ICCV-2017} obtained by removing various components from our framework versus a baseline based on frame-level features and one-class SVM. \label{tab_ablation}}
}
\vspace*{-0.7cm}
\end{table}

\noindent
{\bf Running time.}
The single-shot object detector~\cite{Lin-CVPR-2017} requires about $74$ milliseconds to process a single frame. Hence, it can run at about $13.5$ frames per second (FPS). With a reasonable average of $5$ objects per frame, our feature extraction and inference stages require about $16$ milliseconds per frame. Thus, we can process about $62.5$ frames per second. However, the entire pipeline requires about $90$ milliseconds to infer the anomaly scores for a single frame, which translates to $11$ FPS. We note that more than $80\%$ of the processing time is spent detecting objects on a frame by frame basis. The running time can be improved by replacing the current object detector with a faster one. We note that all running times were measured on an Nvidia Titan Xp GPU with 12 GB of RAM.

\vspace*{-0.2cm}
\section{Conclusion and Future Work}
\label{sec_Conclusion}
\vspace*{-0.1cm}

We introduced a novel method for abnormal event detection in video, which is based on $(i)$ training object-centric convolutional auto-encoders and on $(ii)$ formalizing abnormal event detection as a multi-class problem. The empirical results obtained on four data sets indicate that our approach outperforms a series of state-of-the-art approaches~\cite{Cong-CVPR-2011,Giorno-ECCV-2016,Dutta-AAAI-2015,Hasan-CVPR-2016,Hinami-ICCV-2017,Ionescu-ICCV-2017,Ionescu-WACV-2019,Kim-CVPR-2009,Liu-CVPR-2018,Liu-BMVC-2018,Lu-ICCV-2013,Luo-ICCV-2017,Mahadevan-CVPR-2010,Mehran-CVPR-2009,Ravanbakhsh-WACV-2018,Ravanbakhsh-ICIP-2017,Sabokrou-IP-2017,Saligrama-CVPR-2012,Smeureanu-ICIAP-2017,Sultani-CVPR-2018,Xu-BMVC-2015,Xu-CVIU-2017,Zhang-PR-2016}. In future work, we aim to improve our framework by segmenting and tracking objects.

\noindent
{\bf Acknowledgments.}
The work of Radu Tudor Ionescu was partially supported through grants PN-III-P1-1.1-PD-2016-0787 and PN-III-P2-2.1-PED-2016-1842.

{\small
\bibliographystyle{ieee}
\bibliography{references}
}

\end{document}